\def\year{2020}\relax
\documentclass[letterpaper]{article} 
\usepackage{aaai20}  
\usepackage{times}  
\usepackage{helvet} 
\usepackage{courier}  
\usepackage[hyphens]{url}  
\usepackage{graphicx} 
\usepackage{graphicx}  
\usepackage{amsmath} 
\usepackage{amsfonts} 
\title{Cut-Based Graph Learning networks to Discover Compositional Structure of Sequential Video Data}
\author{Kyoung-Woon On,\textsuperscript{\rm 1} Eun-Sol Kim,\textsuperscript{\rm 2} Yu-Jung Heo,\textsuperscript{\rm 1} Byoung-Tak Zhang\textsuperscript{\rm 1}\\ 
\textsuperscript{\rm 1}Department of Computer Science and Engineering\\ Seoul National University, Seoul, South Korea\\
\textsuperscript{\rm 2}Kakao Brain, Seongnam, South Korea\\
kwon@bi.snu.ac.kr, epsilon.kim@kakaobrain.com, yjheo@bi.snu.ac.kr, btzhang@bi.snu.ac.kr} 

\begin{document}

\maketitle

\begin{abstract}
Conventional sequential learning methods such as Recurrent Neural Networks (RNNs) focus on interactions between consecutive inputs, i.e. first-order Markovian dependency. 
However, most of sequential data, as seen with videos, have complex dependency structures that imply variable-length semantic flows and their compositions, and those are hard to be captured by conventional methods.
Here, we propose Cut-Based Graph Learning Networks (CB-GLNs) for learning video data by discovering these complex structures of the video.
The CB-GLNs represent video data as a graph, with nodes and edges corresponding to frames of the video and their dependencies respectively. 
The CB-GLNs find compositional dependencies of the data in multilevel graph forms via a parameterized kernel with graph-cut and a message passing framework.
We evaluate the proposed method on the two different tasks for video understanding: Video theme classification (Youtube-8M dataset~\cite{abu2016youtube}) and Video Question and Answering (TVQA dataset~\cite{lei2018tvqa}).
The experimental results show that our model efficiently learns the semantic compositional structure of video data. Furthermore, our model achieves the highest performance in comparison to other baseline methods.
\end{abstract}

\section{Introduction}

One of the fundamental problems in learning sequential video data is to find semantic structures underlying sequences for better representation learning.
As most semantic flows cannot be modeled with simple temporal inductive biases, i.e., Markov dependencies, it is crucial to find the complex temporal semantic structures from long sequences to understand the video data. 
We believe there are two ingredients to solving this problem: 
segmenting the whole long-length sequence into (multiple) semantic units and finding their compositional structures.
In this work, we propose a new graph-based method which can discover composite semantic flows in video inputs and utilize them for representation learning of videos. 
The compositional semantic structures are defined as multilevel graph forms, which make it possible to find long-length dependencies and their hierarchical relationships effectively.

In terms of modeling sequential semantic flows, the related work can be summarized with the following three categories: neural networks, clustering, and self-attention based methods. 

In terms of neural network architectures, many problems involving sequential inputs are resolved by using Recurrent Neural Networks (RNNs) as the networks naturally take sequential inputs frame by frame. 
However, as RNN-based methods take frames in (incremental) order, the parameters of methods are trained to capture patterns in transition between successive frames. This makes it hard to find long-term dependencies through overall frames. 
To consider the long-term dependency, Long Short-Term Memory (LSTM)~\cite{hochreiter1997long} and Gated Recurrent Units (GRU)~\cite{chung2014empirical} introduced switches to the RNN structures and Hierarchical RNN~\cite{Chung2017HMRNN} stacked multiple layers to find hierarchical structures.
Even though those methods ignore noisy (unnecessary) frames and maintain the semantic flow through the whole sequence, it is still hard for RNN variants to retain multiple semantic flows and to learn their hierarchical and compositional relationships.
Interestingly, Convolutional Neural Network (CNN) based methods, such as ByteNet~\cite{kalchbrenner2016neural}, ConvS2S~\cite{gehring2017convolutional} and WaveNet~\cite{oord2016wavenet}, applied 1D convolution operator to temporal sequences for modeling dependencies of adjacent frames and their compositions. However these operators hardly captured variable-length dependencies which play a significant role as semantic units.

Recent researches revisited the traditional idea of clustering successive frames into representative clusters.
Deep Bag of Frame (DBoF)~\cite{abu2016youtube} randomly selects $k$ frames from whole sequences as the representatives.
NetVLAD~\cite{arandjelovic2016netvlad} divides all features into $k$ clusters and calculates residuals, which are difference vectors between the feature vectors and their corresponding cluster center, as the new representations for each feature vector. 
Even though the main idea of this type of research is quite simple, it helps to understand the semantics of long sequences by focusing on a small number of representative frames.
However, it is hard to consider the complex temporal relationships.

Along with enormous interest of attention mechanism, a number of significant researches \cite{vaswani2017attention,devlin2019bert} have been aimed to understanding the long (sequential) inputs relying purely on self-attention mechanisms. 
With real-world applications on natural language understanding, such as question and answering (QA) and paragraph detection, those methods focus on meaningful words (or phrases, sentences) among the long reading passages and ignore irrelevant words to the passage by stacking multiple attention layers.
These methods consist of a large number of layers with a huge number of parameters and require a huge amount of training dataset.

In this paper, we propose a method to learn representations of video by discovering the compositional structure in multilevel graph forms.
A single video data input is represented as a graph, where nodes and edges represent frames of the video and relationships between all node pairs.
From the input representations, the Cut-Based Graph Learning Networks (CB-GLNs) find temporal structures in the graphs with two key operations: temporally constrained normalized graph-cut and message-passing on the graph.
A set of semantic units is found by parameterized kernel and cutting operations, then representations of the inputs are updated by message passing operations.

We thoroughly evaluate our method on the large-scale video theme classification task, YouTube-8M dataset~\cite{abu2016youtube}. 
As a qualitative analysis of the proposed model, we visualize compositional semantic dependencies of sequential input frames, which are automatically constructed. 
Quantitatively, the proposed method shows a significant improvement on classification performance over the baseline models.
Furthermore, as an extension of our work, we apply the CB-GLNs to another video understanding task, Video Question and Answering on TVQA dataset~\cite{lei2018tvqa}.
With this experiment, we show how the CB-GLNs can fit into other essential components such as attention mechanism, and demonstrate the effectiveness of the structural representations of our model.

The remainder of the paper is organized as follows. 
As preliminaries, basic concepts of Graph Neural Networks (GNNs) and graph-cut mechanisms are introduced. 
Next, the problem statement of this paper is described to make further discussion clear. 
After that, the proposed Cut-Based Graph Learning Networks (CB-GLNs) are suggested in detail and the experimental results with the real datasets, YouTube-8M and TVQA are presented. 

\section{Preliminaries}\label{sec:pre}

In this section, basic concepts related to graphs are summarized. First, mathematical definitions and notations of graphs are clarified. Second, the normalized graph-cut method is described. Lastly, variants of Graph Neural Networks (GNNs) are introduced.

\subsection{Graph notations}

A graph $G$ is denoted as a pair $(V,E)$ with $V=\{v_1, ..., v_N\}$ the set of nodes (vertices), and $E \in V \times V$ the set of edges. Each node $v_i$ is associated with a feature vector $x_i \in \mathbb{R}^m$. To make notation more compact, the feature matrix of graph $G$ is denoted as $X = [x_1, x_2, ..., x_N]^\top\in\mathbb{R}^{N \times m} $. Also, a graph has an $N$-by-$N$ weighted adjacency matrix $A$ where $A_{i,j}$ represents the weight of the edge between $v_i$ and $v_j$.

\subsection{Normalized graph-cut algorithm}
A graph $G=(V,E)$ can be partitioned into two disjoint sets $V_1$, $V_2$ by removing edges connecting the two parts\footnote{$V_1\cup V_2=V$ and $V_1\cap V_2 = \emptyset $}. 
The partitioning cost is defined as the total weight of the edges that have been removed. 
In addition to the cut cost, the normalized graph-cut method~\cite{shi2000normalized} considers the total edge weight connecting a partition with the entire graph (the degree of the partition) to avoid the trivial solutions which can make extremely imbalanced clusters. 
The objective of the normalized graph-cut can be formally described as follows.

\begin{equation}\label{eq:NCut}
  Ncut(V_1,V_2) = \frac{cut(V_1,V_2)}{assoc(V_1,V)} + \frac{cut(V_1,V_2)}{assoc(V_2,V)}
\end{equation}
with 
\begin{align}
    cut(V_1,V_2) &= \sum_{v_1\in V_1, v_2\in V_2} w(v_1,v_2) \\
  assoc(V_1,V) &= \sum_{v_1\in V_1, v\in V} w(v_1,v)
\end{align}
where $w(v_1,v_2)$ is an edge weight value between node $v_1$ and $v_2$.
It is formulated as a discrete optimization problem and usually relaxed to continuous, which can be solved by eigenvalue problem with the $O(n^2)$ time complexity.
By applying the cut method recursively, an input graph is divided into fine-grained sub-graphs.

\begin{figure*}[ht]
\centering
\includegraphics[width=.95\textwidth]{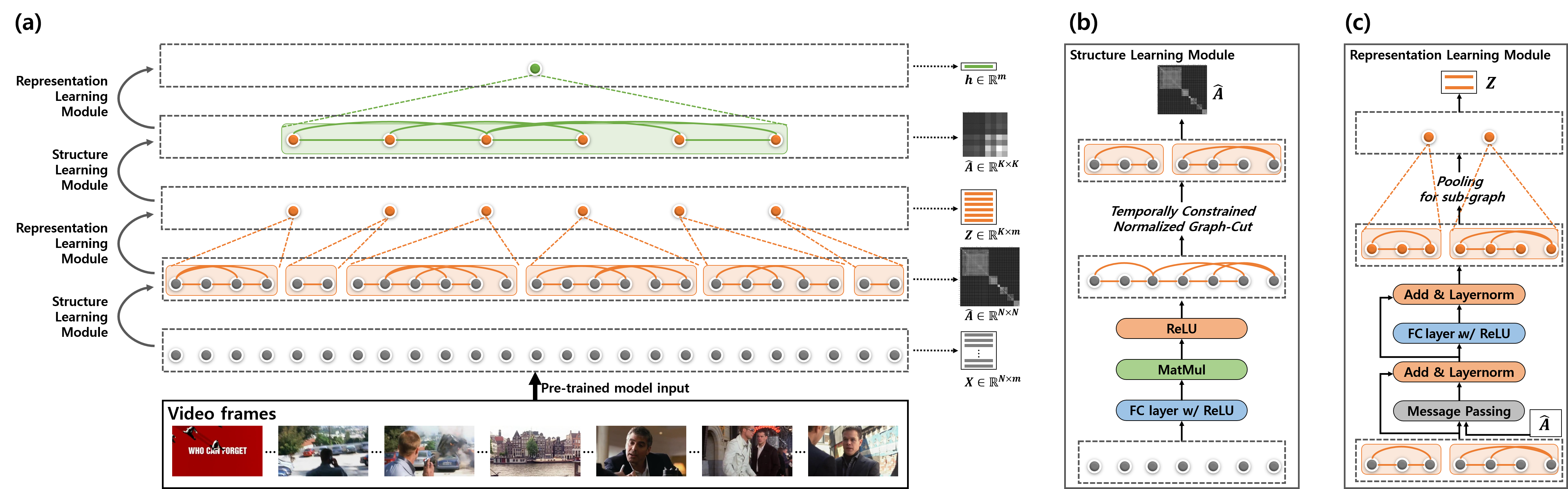}
\caption{(a): The overall architecture of the Cut-Based Graph Learning Networks (CB-GLNs) for a video representation learning task. (b), (c): sophisticated illustrations of Structure Learning Module and Representation Learning Module.}
\label{fig:model_figure}
\end{figure*}

\subsection{Graph Neural Networks (GNNs)}
Since the first proposed by \cite{gori2005new}, interest in combining deep learning and structured approaches has steadily increased. This has led to various graph-based neural networks being proposed over the years.

Based on spectral graph theory~\cite{chung1997spectral}, spectral approaches which convert the graph to the spectral domain and apply the convolution kernel of the graph were proposed~\cite{bruna2013spectral,henaff2015deep,kipf2016semi}.
~\cite{gilmer2017neural} suggested the message passing neural networks (MPNNs), which encompass a number of previous neural models for graphs under a differentiable message passing interpretation.

In detail, the MPNNs have two phases, a message passing phase and a update phase.
In the message passing phase, the message for vertex $v$ in $l$-th layer is defined in terms of message function $M^l$:
\begin{equation}
  m_v^l = \sum_{w\in N(v)} M^l(h_v, h_w)
\end{equation}
where in the sum, $N(v)$ denotes the neighbors of $v$ in a graph $G$.
With the message $m_v^l$, the representations of vertex $v$ in $(l+1)$-th layer are obtained via the update function $U^l$.
\begin{equation}
  h_v^{l+1} = U^l(h_v^l, m_v^l)
\end{equation}
The message functions $M^l$ and the update functions $U^l$ are differentiable so that the MPNNs can be trained in an end-to-end fashion.
As extensions, there are some attempts to have been made to improve message passing by putting a gating or attention mechanism into a message function, which can have computational benefits~\cite{monti2017geometric,duan2017one,hoshen2017vain,velickovic2018graph,garcia2018fewshot,van2018relational}.

We further note other previous research for learning a structure of a graph. Neural Relational Inference (NRI)~\cite{kipf2018neural} used a Variational Autoencoder (VAE)~\cite{kingma2013auto} to infer the connectivity between nodes with latent variables. Other generative models based approaches also have been well studied~\cite{bojchevski2018netgan,de2018molgan,simonovsky2018graphvae}. However, those suffer from availability of structural information in training data or have complex training procedures. To our knowledge, it is the first time to suggest graph-cut based neural networks to discover the inherent structure of videos without supervision of the structural information.

\section{Problem Statement}

The problem to be tackled in this work can be clearly stated with the notations in the previous section as below.

We consider videos as inputs, and a video is represented as a graph $G$.
The graph $G$ has nodes corresponding to each frame respectively in the video with feature vectors
and the dependencies between two nodes are represented with weight values of corresponding edges.

Suppose that video data $X$ has $N$ successive frames 
and each frame has an $m$-dimensional feature vector $x\in \mathbb{R}^{m}$.
Each frame corresponds to a node $v \in V$ of graph $G$, and the dependency between two frames $v_i$, $v_j$ is represented by a weighted edge $e_{ij} \in E$.
From $G=(V,E)$, the dependency structures among video frames are defined as the weighted adjacency matrix $A$, where $A_{ij}=e_{ij}$.
With aforementioned notations and definitions, we can now formally define the problem of video representations learning as follows:

\textit{Given the video frames representations $X\in \mathbb{R}^{N\times m}$, we seek to discover a weighted adjacency matrix $A\in \mathbb{R}^{N\times N}$ which represents dependency among frames.}
\begin{equation}
  f:X\rightarrow A 
\end{equation}
\textit{With $X$ and $A$, final representations for video  $h\in \mathbb{R}^{l}$ are acquired by $g$.}
\begin{equation}
  g:\{X,A\}\rightarrow h 
\end{equation}
The obtained video representations $h$ can be used for various tasks of video understanding.
In this paper, the video theme classification and the video question and answering tasks are mainly considered.

\section{Cut-Based Graph Learning Networks}

The Cut-Based Graph Learning Networks (CB-GLNs) consist of two sub-modules: a structure learning module with the graph-cuts and a representation learning module with message-passing operations.
The key idea of the method is to find inherent semantic structures using the graph-cuts and to learn feature vectors of the video with the message-passing algorithm on the semantic structures.
Stacking these modules leads to the subsequent discovery of compositional structures in the form of a multilevel graph.
Figure \ref{fig:model_figure}(a) illustrates the whole structure of the CB-GLNs.
In the next sections, operations of each of these modules are described in detail.

\subsection{Structure Learning Module} \label{sec:SLL}
In the structure learning module, the dependencies between frames $\hat{A}$ are estimated via parameterized kernels and the temporally constrained graph-cut algorithm.

As the first step, the initial temporal dependencies over all frames are constructed via the parameterized kernel $\mathcal{K}$:
\begin{equation}\label{eq:kernel}
  \hat{A}_{ij} = \mathcal{K}(x_i,x_j) = ReLU(f(x_i)^\top f(x_j))
\end{equation}
where $f(x)$ is a single-layer feed-forward network without non-linear activation.

Then, as the second step, the meaningful dependency structure among all pairwise relationships is refined by applying normalized graph-cut algorithm to the $\hat{A}$. 
The objective of the normalized graph-cut for CB-GLNs is:
\begin{equation}
  Ncut(V_1,V_2)\!=\! \frac{\sum_{v_i\in V_1, v_j\in V_2} \!\hat{A}_{ij}}{\sum_{v_i\in V_1} \hat{A}_{i\cdot }}\! +\! \frac{\sum_{v_i\in V_1, v_j\in V_2} \hat{A}_{ij}}{ \sum_{v_j\in V_2}\! \hat{A}_{j\cdot}}
\end{equation}\label{eq:ncut-cb-gln}

To reduce the complexity of the equation (9)\ref{eq:ncut-cb-gln} and to keep the inherent characteristics of the video data, an additional constraint is added to the graph-cut algorithm.
As the video data is composed of time continuous sub-sequences, no two partitioned sub-graphs have an overlap in physical time. This characteristic is implemented by applying the temporal constraint~ \cite{rasheed2005detection,sakarya2008graph} as follows.

\begin{equation}\label{eq:NCut_Const}
  (i<j\; \text{or}\; i>j)\;\; \text{for all}\;v_{i}\in V_1, v_j\in V_2
\end{equation}
Thus, a cut can only be made along the temporal axis and complexity of the graph partitioning is reduced to linear time while keeping the characteristics of the video data.
Also, as the gradients can flow through the surviving edges, it is end-to-end trainable.

The graph-cut can be recursively applied to the $\hat{A}$, so  $K$ partitioned sub-graphs can be obtained.
The number of cut operations is determined by the length of the video $N$, $\lfloor\log_{2}\sqrt{N}\rfloor$, and we also add the constraint that sub-cluster should not be partitioned if it is no longer than pre-specified length. 
Figure \ref{fig:model_figure}(b) depicts the detailed operations of the structure learning module.

\subsection{Representation Learning Module}

After estimating the weighted adjacency matrix $\hat{A}$, the representation learning module updates the representations of each frame via a differentiable message-passing framework~\cite{gilmer2017neural}. 
For the message function $F_M$, we simply use the weighted sum of adjacent nodes' representations after linear transformation similar to \cite{kipf2016semi}:
\begin{align}
  M=F_M(X,\hat{A}_{\text{cut}}) &=  D^{-1}\hat{A}_{\text{cut}}XW_M 
\end{align}
where $D$ is a degree matrix of the graph and $\hat{A}_{\text{cut}}$ is an adjacency matrix after cut operations.

For the update function $F_U$, we integrate the message with node representations by using low-rank bilinear pooling $B$~\cite{DBLP:conf/iclr/KimOLKHZ17} followed by a position-wise fully connected network.
\begin{align}
  H = F_U(X,M) &=  f(B(X,M))
\end{align}
where $f$ is a single-layer position-wise fully connected network.
We also employ a residual connection~\cite{he2016deep} around each layer followed by layer normalization~\cite{ba2016layer}.

Once the representations of all frames are updated, a pooling operation for each partitioned sub-graph is applied.
Then we can obtain higher level representations $Z\in\mathbb{R}^{K\times m}$, where $K$ is the number of partitioned sub-graphs (Figure \ref{fig:model_figure}(c)). 
If we have additional information such as query (e.g. a question feature vector in video QA setting), we can pool the sub-graph with attentive pooling similar to ~\cite{santos2016attentive}.

In the same way, $Z$ is fed into the new structure learning module and we can get the video-level representation $h\in \mathbb{R}^m$.

\section{Experiments}
In this section, the experimental results on the two different video datasets, YouTube-8M (video theme classification) and TVQA (video question and answering), are provided.

\subsection{Video Theme Classification Task on YouTube-8M}
\subsubsection{Data specification}
YouTube-8M~\cite{abu2016youtube} is a benchmark dataset for video understanding, where the main task is to determine the key topical themes of a video.
The dataset consists of 6.1M video clips collected from YouTube.
Each video is labeled with one or multiple tags referring to the main topic of the video. Each video is encoded at 1 frame-per-second up to the first 300 seconds.
The volume of video data is too large to be treated in it's raw form. As such the input is pre-processed with pretrained models by the author of the dataset.

Global Average Precision (GAP) is used for the evaluation metric for the multi-label classification task as used in the YouTube-8M competition. For each video, 20 labels are predicted with confidence scores, then the GAP score computes the average precision and recall across all of the predictions and all the videos. 

\subsubsection{Model setup}
The frame-level visual and audio features are extracted by inception-v3 network~\cite{szegedy2016rethinking} trained on imagenet and VGG-inspired architecture~\cite{hershey2017cnn} trained for audio classification. 
These features construct an input feature matrix $X$ of video sequences, then $X$ is fed into a sequence model to extract final video representation $h$.
Including our model, all baseline sequence models are composed of two layers and average pooling is used for final representations.
With the representation $h$, a simple logistic regression is used as a final classifier.

\subsubsection{Quantitative results}

\begin{table}\label{table:exp_1}
\begin{center}
\caption{Comparison on classification accuracy with the GAP measure on validation dataset. Logistic regression is as a classifier for all of the presented methods.}
\begin{tabular}{ll}
\hline\noalign{\smallskip}
Frame-level model & GAP \\
\noalign{\smallskip}
\hline
Average pooling   & 0.7824 \\
\hline
DeepBoF (4096 clusters)   & 0.8079 \\
NetVLAD (256 clusters)  & 0.8396 \\
\hline
1D CNN (2 layers, kernel size 3) & 0.8254  \\
1D CNN (2 layers, kernel size 5) & 0.8245  \\
1D CNN (2 layers, kernel size 7) & 0.8247 \\
\hline
LSTM (2 layers)   & 0.8446 \\
GRU (2 layers)   & 0.8160\\
BiLSTM (2 layers)   & 0.8410\\
BiGRU (2 layers)  & 0.8079\\
\hline
Self-Attention (4 heads, 2 layers) & 0.8553 \\
\hline
NeXtVLAD~\cite{lin2018nextvlad} & 0.8499 \\
DCGN~\cite{mao2018hierarchical} & 0.8450 \\
\hline
\textbf{CB-GLNs (2 layers)} & \textbf{0.8597}  \\
\hline
\end{tabular}
\end{center}
\end{table}
\setlength{\tabcolsep}{8pt}

Firstly, we evaluate the classification performance of the proposed model against four types of representative sequential models described in the Introduction section and two state-of-the-art models~\cite{lin2018nextvlad,mao2018hierarchical} previously reported.

The results with GAP score are summarized in Table 1.
The proposed model considerably outperforms all the comparative models. The second best performing model is the self-attention based approach, followed by  RNNs, CNNs and Clustering based approaches.
In the ``Qualitative results" Section, automatically constructed compositional structures are discussed for better understanding of the model.

For ablation studies, we selected three critical characteristics of CB-GLNs for in-depth analysis: layer normalization, residual connection and graph-cut after learning representations. 
The GAP scores on the validation set for each ablation experiments are shown in Table 2.
As can be seen from (a) to (c) in the Table 2, the residual connections followed by layer normalization are crucial for the representation learning module.
Also, to see the effect of sparsening an adjacency matrix via the graph-cut, reversed order of representation learning and graph-cut is also conducted ((d) in Table 2). By doing so, the representations of each node are updated with $\hat{A}$ obtained only using kernel $\mathcal{K}$ (in Equation \ref{eq:kernel}) and graph-cut algorithm is used just for sub-graph pooling.
Thus, the model has to learn representations with dense and noisy connections, degrading the performance of the model. 
From this result, we can argue that the temporally constrained graph-cut effectively reduce the noisy connections in the graph.

\setlength{\tabcolsep}{4pt}
\begin{table}\label{table:exp_2}
\begin{center}
\caption{An ablation study of Cut-Based Graph Learning Networks.}
\begin{tabular}{ll}
\hline\noalign{\smallskip}
Ablation model & GAP \\
\noalign{\smallskip}

\hline
(a) layer normalization   & 0.8486 \\
(b) residual connection   & 0.8447 \\
(c) residual connection with layer normalization & 0.8370\\
(d) graph-cut after learning representations & 0.8576\\
\noalign{\smallskip}
\hline
CB-GLNs   & 0.8597 \\
\hline
\end{tabular}
\end{center}
\end{table}
\setlength{\tabcolsep}{1.4pt}

\subsubsection{Qualitative results: Learning compositional dependency structure}\label{sec:qr}

\begin{figure*}[t] 
\centering
\includegraphics[width=.90\linewidth]{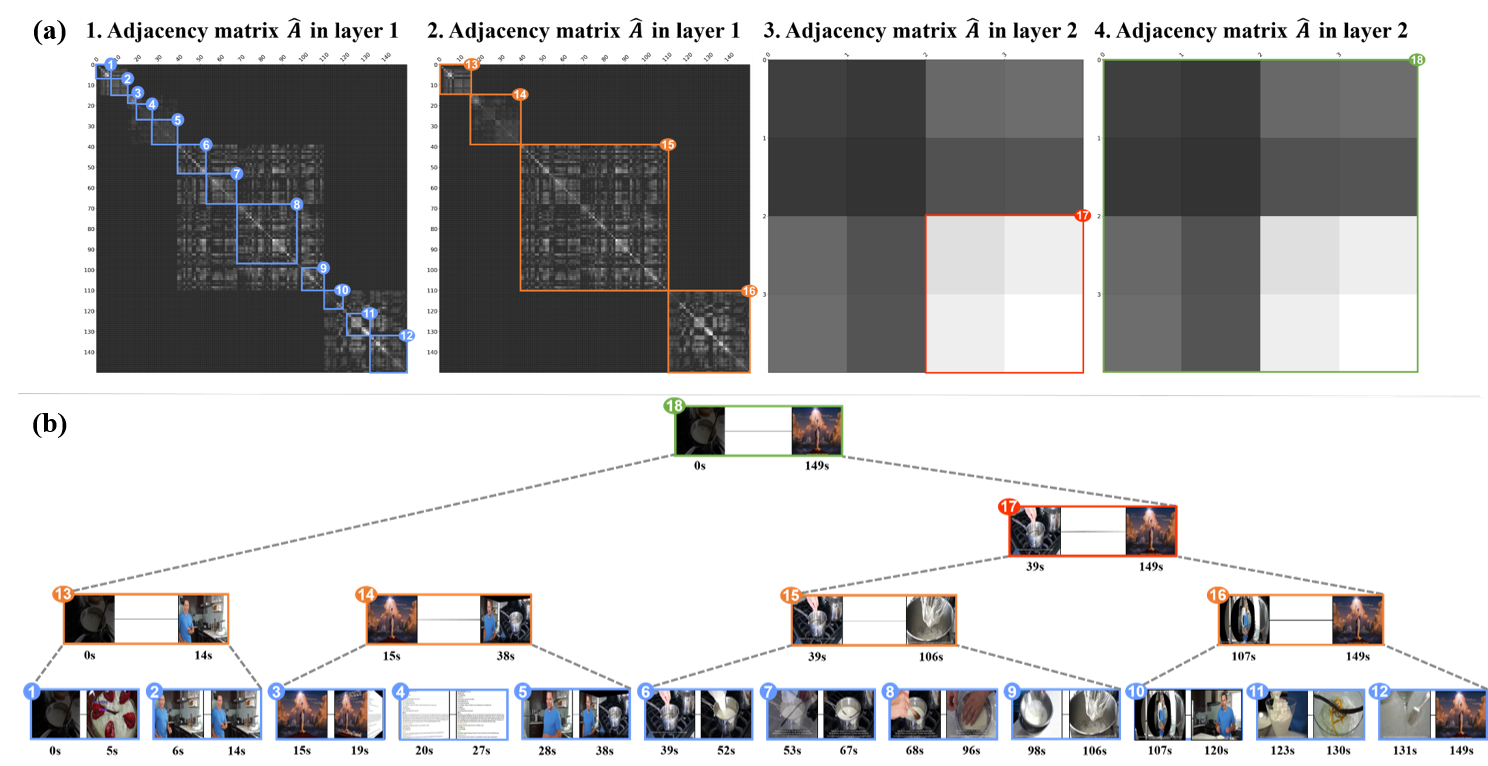}

\caption{An example of the constructed temporal dependency structures for a real input video in YouTube-8M, titled ``Rice Pudding" (https://youtu.be/cD3enxnS-JY) is visualized. The topical themes (true labels) of this video are \{Food, Recipe, Cooking, Dish, Dessert, Cake, baking, Cream, Milk, Pudding and Risotto\}. (a): Learned adjacency matrices in the layer 1 and 2 are visualized. The strength of connections are encoded in a gray-scale where 1 is white and 0 is black. (a)-1: 12 bright blocks in layer 1 are detected (blue rectangles), each block (highly connected frames) represents a semantic unit. (a)-2: Sub-graphs of the input are denoted by orange rectangles. It shows that semantically meaningful scenes are found by temporally constrained graph-cut.  (a)-3 and (a)-4: learned high-level dependency structures in layer 2 are revealed with red and green rectangles. 
(b): The whole composite temporal dependencies are presented.}
\vspace{-0.5cm}
\label{fig:qualitative_result} 
\end{figure*}

In this section, we demonstrate compositional learning capability of CB-GLNs by analyzing constructed multilevel graphs.
To make further discussion clear, four terms are used to describe the compositional semantic flows: semantic units, scenes, sequences and a video for each level.
In Figure \ref{fig:qualitative_result}, a real example with the usage of video titled ``Rice Pudding\footnote{https://youtu.be/cD3enxnS-JY}" is described to show the results. 

In Figure \ref{fig:qualitative_result}(a), the learned adjacency matrices in each layer are visualized in gray-scale images: the two leftmost images are from the 1st layer and the two rightmost images are from the 2nd layer. To denote multilevel semantic flows, four color-coded rectangles (blue, orange, red and green) are marked and those colors are consistent with Figure \ref{fig:qualitative_result}(b).

Along with diagonal elements of the adjacency matrix in the 1st layer (Figure \ref{fig:qualitative_result}(a)-1), a set of semantic units are detected corresponding to bright blocks (blue). 
Interestingly, we found that each semantic unit contains highly correlated frames.
For example, the \#1 and \#2 are each shots introducing the YouTube cooking channel and how to make rice pudding, respectively. The \#4 and \#5 are shots showing a recipe of rice pudding and explaining about the various kinds of rice pudding. The \#6 and \#7 are shots putting ingredients into boiling water in the pot and bringing milk to boil along with other ingredients. At the end of the video clip, \#11 is a shot decorating cooked rice pudding and \#12 is an outro shot that invites the viewers to subscribe.

These semantic units compose variable-length scenes of the video, and each scene corresponds to a sub-graph obtained via graph-cut (Figure \ref{fig:qualitative_result}(a)-2.). 
For example, \#13 is a scene introducing this cooking channel and rice pudding. Also, \#15 is a scene of making rice pudding with detailed step by step instructions, and \#16 is an outro scene wrapping up with cooked rice pudding. 
The 1st-layer of the model updates representations of frame-level nodes with these dependency structures, then aggregates frame-level nodes to form scene-level nodes (Layer 1 in the Figure \ref{fig:qualitative_result}(b)).

In Figure \ref{fig:qualitative_result}(a)-3 and (a)-4, the sequence-level semantic dependencies (red) are shown.  \#17 denotes a sequence of making rice pudding from beginning to end, which contains much of the information for identifying the topical theme of this video. Finally, the representations of scenes are updated and aggregated to get representations of the whole video (Layer 2 in the Figure\ref{fig:qualitative_result}(b)).

\subsection{Video Question \& Answering Task on TVQA}
\subsubsection{Data specification}
TVQA~\cite{lei2018tvqa} is a video question and answering dataset on TV show domain. It consists of total 152.5k question-answer pairs on six TV shows: The Big Bang Theory, How I Met Your Mother, Friends, Grey’s Anatomy, House and Castle. Also, it contains 21.8k short clips of 60-90 seconds segmented from the original TV show for question-answering.
The provided inputs are 3 fps image frames, subtitles and multiple choice questions with 5 candidate answers for each question, for which only one is correct.
 
The questions in the dataset are localized to a specific sub-part in the video clips by restricting questions to a composition of two parts, e.g., "Where was Sheldon sitting / before he spilled the milk?".
Models should answer questions using both visual information and associated subtitles from the video.

\subsubsection{Model setup}
The input visual features were extracted by the pooled 2048D feature of the last block of  ResNet101~\cite{he2016deep} trained on imagenet and the text-based features for subtitles, questions and answers were extracted by GloVe~\cite{pennington2014glove}.
The visual features and subtitle features are manually aligned with time-stamp and answer features also aligned with visual and subtitle features by attention mechanism to construct input $X$.
Then the $X$ is fed into the CB-GLNs to extract final representations $h$.
Different from the YouTube-8M dataset case, we use a question feature vector as a query of attentive pooling, so that the representations of the sequence are pooled via weighted sum with the attention values.

\subsubsection{Qualitative results: Attention hierarchy with learned compositional structure}

\begin{figure*}[t] 
\centering
\includegraphics[width=.90\linewidth]{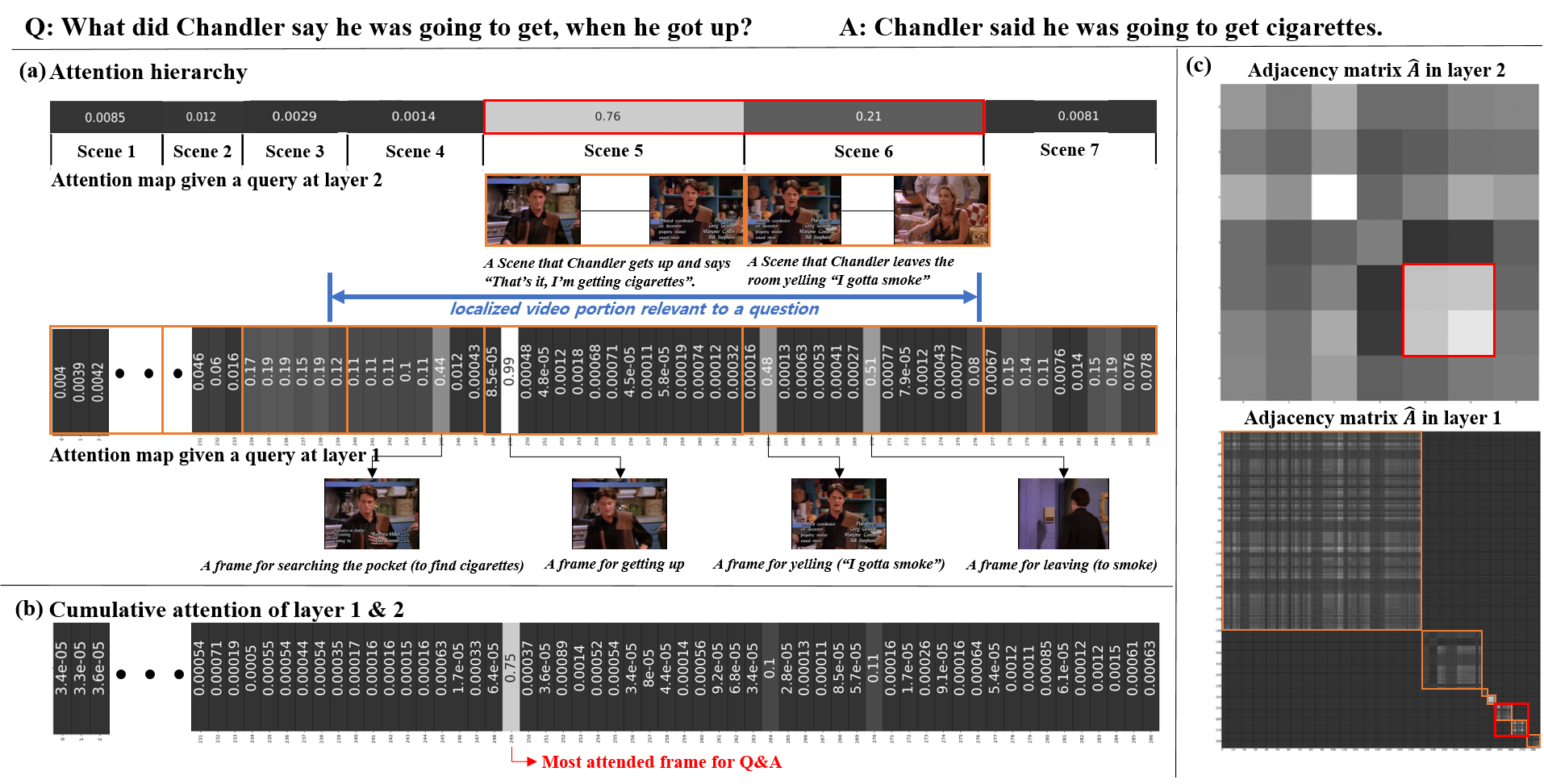}

\caption{An example of the learned attention hierarchy for a real video clip of ``Friends" in TVQA is visualized. The question and answer for this clip are ``What did Chandler say he was going to get, when he got up?" and ``Chandler said he was going to get cigarettes." each. We dropped the Scene 1 and Scene 2 parts in (a) and (b), because the attention values in these scenes are nearly identical, making them non-informative.
(a): Attention maps for each layer given query. the orange rectangles in layer 1 denote the cut scenes discovered by CB-GLNs and the red rectangle in layer 2 highlights scenes with high attention values given a question. (b): Cumulative attentions to the frame-level are visualized by multiplying attention values in layer 1 \& 2. (c): Visualization of the learned adjacency matrices in layer 1 \& 2. The orange and red rectangles are consistent with (a).}
\label{fig:qualitative_result2}
\end{figure*}

In this section, we show how the attention mechanism can be fit into the CB-GLNs to learn representations more effectively.
Basically, the attention mechanism places more weight on important parts to aggregate values, given a query.
By virtue of the compositionality, the attention mechanism can be naturally applied to the CB-GLNs in a hierarchical fashion.
Figure~\ref{fig:qualitative_result2} presents learned attention hierarchy in a real video clip of ``Friends". In this example, The question is ``What did Chandler say he was going to get, when he got up?" and the answer for the question is that ``Chandler said he was going to get cigarettes.". 

In Figure~\ref{fig:qualitative_result2}(a), we can see that scenes with high attention values (Scene 5 and Scene 6 (coded by a red rectangle)) in layer 2 are aligned well with localized video portions relevant to a given question.
Scene 4, where ``Chandler is searching his pocket to find cigarettes while sitting down", is also in the localized section. Therefore, we can say our model finds a sensitive portion of the video relevant to the given question.
In layer 1 (Figure~\ref{fig:qualitative_result2}(a)), the model gives a keener attention to the frame-level within each scenes (coded by orange rectangles), such as a moment where Chandler gets up or Chandler yells `I gotta smoke`.

Because the attention operation for each frame is conducted hierarchically, we can calculate cumulative attention values by multiplying them in layer 1 and layer 2.
Figure~\ref{fig:qualitative_result2}(b) shows the cumulative attention values for each frame. 
The most important frame in a viewpoint of the model is the "getting up moment" frame because the model should answer the question by identifying the meaning of ``when he got up" in the question.

In Figure~\ref{fig:qualitative_result2}(c), the learned adjacency matrices in each layer are visualized. Same color-coded rectangles with (a) are used for cut scenes (orange rectangles in layer 1) and scenes with high attention values (a red rectangle in layer 2). 
We also coded red rectangle in layer 1, which is corresponding to scenes with high attention values in layer 2.
Even though scene 5 and scene 6 in the frame level (layer 1) are considerably short when compared to the whole video sequences, the CB-GLNs can find important moments and aggregate them in an effective way.

\section{Conclusion}
In this paper, we proposed Cut-Based Graph Learning Networks (CB-GLNs) which learn not only the representations of video sequences, but also composite dependency structures within the sequence. To explore characteristics of CB-GLNs, various experiments are conducted on a real large-scale video dataset YouTube-8M and TVQA. The results show that the proposed model efficiently learns the representations of sequential video data by discovering inherent dependency structure of itself.

\section{Acknowledgements}
The authors would like to thank Woo Suk Choi and Chris Hickey for helpful comments and editing. This work was partly supported by the Korea government (2015-0-00310-SW.StarLab, 2017-0-01772-VTT, 2018-0-00622-RMI, 2019-0-01367-BabyMind, P0006720-GENKO).

\bibliographystyle{aaai20}
\bibliography{AAAI-OnK.7474}

\end{document}